\documentclass{article}
\usepackage{spconf,amsmath,graphicx}

\usepackage{url}
\usepackage[hidelinks]{hyperref}
\usepackage[utf8]{inputenc}
\usepackage[small]{caption}
\usepackage{amsthm}
\usepackage{bbm}
\usepackage{booktabs}
\usepackage{algorithm}
\usepackage{algorithmic}
\usepackage{multirow}
\DeclareMathOperator*{\argmax}{arg\,max}

\usepackage{tikz}

\title{Speaker Diarization as a Fully Online Learning Problem in MiniVox}
\name{
Baihan Lin$^{1}$,
Xinxin Zhang$^{2}$}
\address{
$^1$ Columbia University\\
$^2$ New York University\\
\tt{baihan.lin@columbia.edu, xz3149@nyu.edu}
}

\begin{document}
%
\maketitle

\begin{abstract}

We proposed a novel machine learning framework to conduct real-time multi-speaker diarization and recognition without prior registration and pretraining in a fully online learning setting. Our contributions are two-fold. First, we proposed a new benchmark to evaluate the rarely studied fully online speaker diarization problem. We built upon existing datasets of real world utterances to automatically curate \textit{MiniVox}, an experimental environment which generates infinite configurations of continuous multi-speaker speech stream. Second, we considered the practical problem of online learning with episodically revealed rewards and introduced a solution based on semi-supervised and self-supervised learning methods. Additionally, we provided a workable web-based recognition system which interactively handles the cold start problem of new user's addition by transferring representations of old arms to new ones with an extendable contextual bandit. We demonstrated that our proposed method obtained robust performance in the online MiniVox framework.
\footnote{The web-based application of a real-time system can be accessed at \url{https://www.baihan.nyc/viz/VoiceID/} (as in \cite{lin2020voiceid}). The code for benchmark evaluation can be accessed at \url{https://github.com/doerlbh/MiniVox}.}
\end{abstract}


\section{Introduction}
\vspace{-0.1cm}

Speaker recognition involves 
two essential steps: registration and identification \cite{tirumala2017speaker}. 
In laboratory setting, the state-of-the-art approaches usually emphasize the registration step with deep networks \cite{snyder2018x} trained on large-scale speaker profile dataset \cite{Nagrani17}. However, in real life, requiring all users to complete voiceprint registration before a multi-speaker teleconference is hardly a preferable way of system deployment. Dealing with this challenge, speaker diarization is the task to partition an audio stream into homogeneous segments according to the speaker identity \cite{anguera2012speaker}. Recent advancements have enabled (1) contrastive audio embedding extractions such as Mel Frequency Cepstral Coefficients (MFCC) \cite{hasan2004speaker}, i-vectors \cite{shum2013unsupervised} and d-vectors \cite{wang2018speaker}; (2) effective clustering modules such as Gaussian mixture models (GMM) \cite{zajic2017speaker}, mean shift \cite{senoussaoui2013study}, Kmeans and spectral clustering \cite{wang2018speaker} and supervised Bayesian non-parametric methods \cite{fox2011sticky,zhang2019fully}; and (3) reasonable resegmentation modules such as Viterbi and factor analysis subspace \cite{sell2015diarization}. In this work, we proposed a new paradigm to consider the speaker diarization as a fully online learning problem of the speaker recognition task: it combines the embedding extraction, clustering and resegmentation into the same problem as an online decision making problem.

\textbf{Why is this online learning problem different?} The state-of-the-art speaker diarization systems usually require large datasets to train their audio extraction embeddings and clustering modules, especially the ones with deep neural networks and Bayesian nonparametric models. In many real-world applications in developing countries, however, the training set can be limited and hard to collect. Since these modules are pretrained, applying them to out-of-distribution environments can be problematic. For instance, an intelligent system trained with American elder speaker data might find it hard to generalize to a Japanese children diarization task because both the acoustic and contrastive features are different. To tackle this problem, we want the system to learn continually. To push this problem to the extreme, we are interested in a fully online learning setting, where not only the examples are available one by one, the agent receives no pretraining from any training set before deployment, and learns to detect speaker identity on the fly through reward feedbacks. To the best of our knowledge, this work is the first to consider diarization as a fully online learning problem. Through this work, we aim to understand the extent to which diarization can be solved as merely an online learning problem and whether traditional online learning algorithms \hfill (e.g. contextual bandits \cite{langford2007epoch,lin2020unified,lin2020ipd}) can be a practical solution.

\textbf{What is a preferable online speaker diarization system?} A preferable AI engine for such a realistic speaker recognition and diarization system should (1) not require user registrations before its deployment, (2) allow new user to be registered into the system real-time, (3) transfer voiceprint information from old users to new ones, (4) be up running without pretraining on large amount of data in advance. While attractive, assumption (4) introduced an additional caveat that the labeling of the user profiles happens purely on the fly, trading off models pretrained on big data with the user directly interacting with the system by correcting the agent as labels. To tackle these challenges, we formulated this problem into an interactive learning model with cold-start arms and episodically revealed rewards (users can either reveal no feedback, approving by not intervening, or correcting the agent). 

\textbf{Why do we need a new benchmark?} Traditional dataset in the speaker diarization task are limited: CALLHOME American English \cite{canavan1997callhome} and NIST RT-03 English CTS \cite{martin2000nist} contained limited number of utterances recorded under controlled conditions. For online learning experiments, a learn-from-scratch agent usually needs a large length of data stream to reach a comparable result. Large scale speaker recognition dataset like VoxCeleb \cite{Nagrani17,Nagrani19} and Speakers in the Wild (SITW) \cite{mclaren2016speakers} contained thousands of speaker utterances recorded in various challenging multi-speaker acoustic environments, but they are usually only used to pretrain diarization embeddings. In this work, we proposed a new benchmark called \textit{MiniVox}, which can transform any large scale speaker identification dataset into infinitely long online audio streams with various configurations.  

To the best of our knowledge, this is the first approach to apply the \textit{Bandit} problem to the speaker diarization task. We built upon the Linear Upper Confidence Bound algorithm (LinUCB) \cite{li2010contextual} and proposed a semi-supervised learning variant to account for the fact that the rewards are entirely missing in many episodes. For each episode without feedbacks, we applied a self-supervision process to assign a pseudo-action upon which the reward mapping is updated. Finally, we generated new arms by transferring learned arm parameters to similar profiles given user feedbacks. 




\begin{figure}[tb]
\centering
   \includegraphics[width=0.75\linewidth]{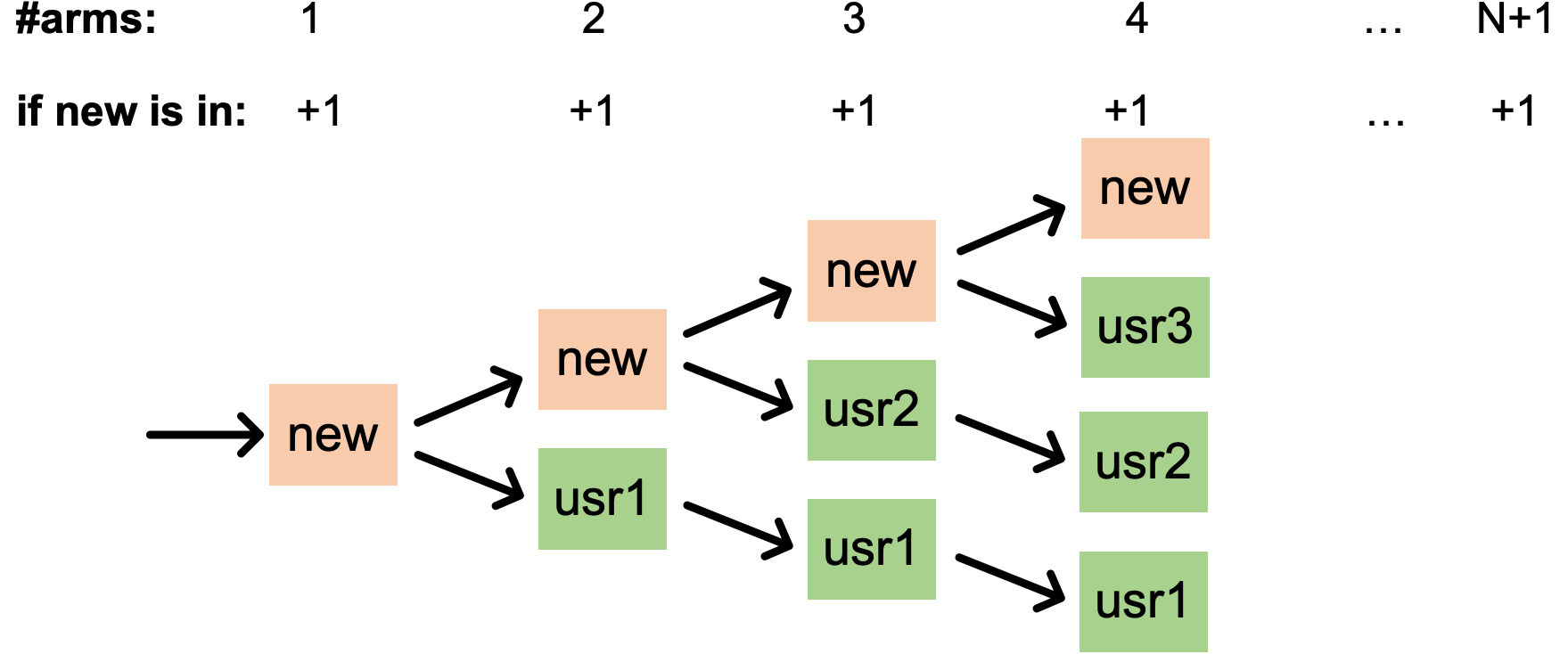}
\par\caption{The arm expansion process of the bandit agents.}\label{fig:bandit}
\vspace{-0.2cm}
\end{figure}

\vspace{-0.2cm}
\section{Background: the Bandit Problem}
\vspace{-0.1cm}

In the online learning setting, the data become available in a sequential order and later used to update the best predictor for future data or reward associated with the data features. In many case, the reward feedback is the only source where the online learning agent can effectively learn from the sequential past experience. This problem is especially important in the field of sequential decision making where the agent must choose the best possible action to perform at each step to maximize the cumulative reward over time. One key challenge is to obtain an optimal trade-off between the exploration of new actions and the exploitation of the possible reward mapping from known actions. This framework is usually formulated as the \textit{Bandit} problem where each arm of the bandit corresponds to an unknown (but usually fixed) reward probability distribution \cite{LR85}, and the agent selects an arm to play at each round, receives a reward feedback and updates accordingly. An especially useful variant of Bandit is the \textit{Contextual Bandit}, where at each step, the agent observes an $N$-dimensional \textit{context}, or \textit{feature} vector before selecting an action. Theoretically, the ultimate goal of the Contextual Bandit is to learn the relationship between the rewards and the context vectors so as to make better decisions given the context \cite{AgrawalG13}.

\vspace{-0.2cm}
\section{The Fully Online Learning Problem}
\vspace{-0.1cm}

Algorithm \ref{alg:episode} presents at a high-level our problem setting of our interactive learning system for speaker diarization, where $x(t)\in \mathbbm{R}^d$ is a vector describing the context $C$ at time $t$, $r_{a,t}(t) \in [0,1]$ is the reward of action $a$ at time $t$, and $r(t) \in [0,1]^N$ denotes a vector of rewards for all arms at time $t$. $\mathbbm{P}_{x,r}$ denotes a joint probability distribution over $(x,r)$, and $\pi: C \rightarrow A$ denotes a policy. Unlike traditional setting, in step 5 we have the rewards revealed in an episodic fashion (i.e. sometimes there are feedbacks of rewards being 0 or 1, sometimes there are no feedbacks of any kind). We consider our setting online semi-supervised learning \cite{Yver2009}, where agents continually learn from both labeled and unlabeled data.

\begin{algorithm}[tb]
\small
 \caption{Online Learning with Episodic Rewards}
 \label{alg:episode}
 \begin{algorithmic}[1]
 \STATE {\bfseries }\textbf{for} t = 1,2,3,$\cdots$, T \textbf{do}
\STATE {\bfseries } \quad $(\textbf{x}(t),\textbf{r}(t))$ is drawn according to $\mathbbm{P}_{x,r}$
\STATE {\bfseries }\quad  Context $\textbf{x}(t)$ is revealed to the player
\STATE {\bfseries }\quad  Player chooses an action $a_t =\pi_t(\textbf{x}(t))$
\STATE {\bfseries } \quad Feedback $r_{a_t,t}(t)$ for the arm $a_t$ is episodically revealed
\STATE {\bfseries } \quad Player updates its policy $\pi_t$
\STATE {\bfseries } \textbf{end for}
 \end{algorithmic}
\end{algorithm}
\setlength{\textfloatsep}{0.45cm}
\setlength{\floatsep}{0.45cm}


\begin{figure*}[tb]
\centering
    \includegraphics[width=0.91\linewidth]{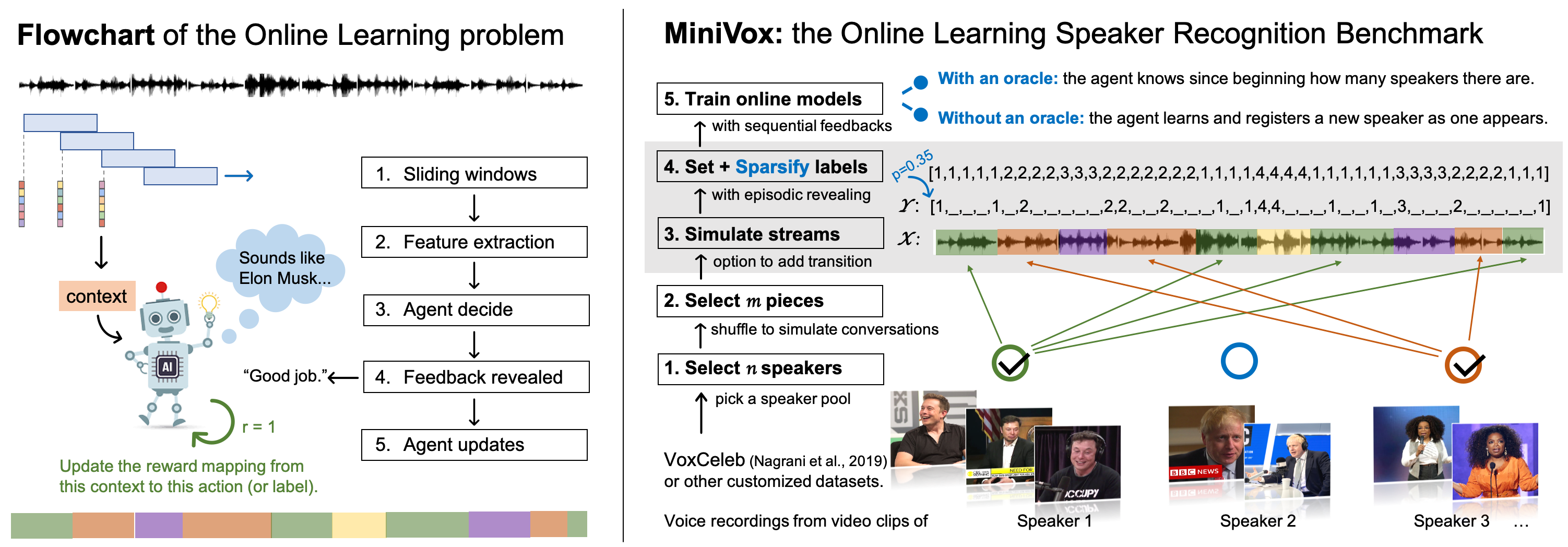}
\vspace{-0.1cm}
\caption{(A) The \textbf{flowchart} of the speaker diarization task as a Fully Online Learning problem and (B) the \textbf{\textit{MiniVox} Benchmark}. 
}\label{fig:minivox}
\vspace{-0.3cm}
\end{figure*}

\vspace{-0.2cm}
\section{Proposed Online Learning Solution}
\vspace{-0.2cm}

\subsection{Contextual Bandits with Extendable Arms}
\label{sec:banditarm}
\vspace{-0.1cm}

In an ideal online learning scenario without oracle, we start with a single arm, and when new labels arrive new arms are then generated accordingly. This problem is loosely modelled by the bandits with infinitely many arms \cite{berry1997bandit}. For our specific application of speaker registration process, we applied the arm expansion process outlined in Figure \ref{fig:bandit}: starting from a single arm (for the ``new'' action), if a feedback confirms a new addition, a new arm is initialized and stored (more details on the handling of growing arms can be found in section \ref{sec:engine}).

\vspace{-0.2cm}

\subsection{Episodically Rewarded LinUCB}
\vspace{-0.1cm}

We proposed Background Episodically Rewarded LinUCB (BerlinUCB
\cite{lin2020online}), 
a semi-supervised and self-supervised online contextual bandit which updates the context representations and reward mapping separately given the state of the feedbacks being present or missing. As in Algorithm \ref{alg:berlinucb}, the steps 1 through 12 of BerlinUCB are the same as the standard LinUCB algorithm \cite{li2010contextual}, and in case of a missing reward, we introduced the steps 13 through 20 as the alternative strategy. We assume that: (1) when there are feedbacks available, the feedbacks are genuine, assigned by the oracle, and (2) when the feedbacks are missing (not revealed by the background), it is either due to the fact that the action is preferred (no intervention required by the oracle, i.e. with an implied default rewards), or that the oracle didn't have a chance to respond or intervene (i.e. with unknown rewards). Especially in the Step 15, when there is no feedbacks, we assign the context $\textbf{x}_t$ to a class $a'$ (an action arm) with the self-supervision given the previous labelled context history (section \ref{sec:clustering}). Since we don't have the actual label for this context, we only update the reward mapping parameter $\textbf{b}_{a'}$ and leave the covariance matrix $\textbf{A}_{a'}$ untouched. This additional usage of unlabelled data (or unrevealed feedback) is especially important in our model. 

\begin{algorithm}[tb]
\small
 \caption{BerlinUCB}
 \label{alg:berlinucb}
 \begin{algorithmic}[1]
 \STATE {\bfseries } \textbf{Initialize} $c_t \in \mathbbm{R}_+, \textbf{A}_a \leftarrow \textbf{I}_d, \textbf{b}_a \leftarrow \textbf{0}_{d \times 1} \forall a \in \mathcal{A}_t $
 \STATE {\bfseries }\textbf{for} t = 1,2,3,$\cdots$, T \textbf{do}
  \STATE {\bfseries } \quad Observe features $\textbf{x}_{t}\in\mathbbm{R}^d$
 \STATE {\bfseries } \quad \textbf{for all} $a \in \mathcal{A}_t $ \textbf{do}
 \STATE {\bfseries }  \quad \quad  $\hat{\mathbf{\theta}}_a \leftarrow \textbf{A}_a^{-1}\textbf{b}_a$
 \STATE {\bfseries } \quad \quad $p_{t,a} \leftarrow \hat{\mathbf{\theta}}_a^{\top}\textbf{x}_{t}+c_t\sqrt{\textbf{x}_{t}^{\top}\textbf{A}_a^{-1}\textbf{x}_{t}}$
\STATE {\bfseries }  \quad \textbf{end for}
\STATE {\bfseries } \quad Choose arm $a_t=\argmax_{a \in \mathcal{A}_t}p_{t,a}$ 
\STATE {\bfseries } \quad \textbf{if} the background revealed the feedbacks \textbf{then} 
 \STATE {\bfseries }  \quad \quad Observe feedback $r_{a_t,t}$ 
\STATE {\bfseries }  \quad \quad $\textbf{A}_{a_t} \leftarrow \textbf{A}_{a_t}+\textbf{x}_{t}\textbf{x}^{\top}_{t}$ 
\STATE {\bfseries }  \quad \quad $\textbf{b}_{a_t} \leftarrow \textbf{b}_{a_t}+r_{a_t,t}\textbf{x}_{t}$ 
\STATE {\bfseries } \quad \textbf{elif} the background revealed NO feedbacks \textbf{then} 
\STATE {\bfseries } \quad \quad  \textbf{if} use self-supervision feedback
\STATE {\bfseries } \quad \quad  \quad $r' = [a_t == \text{predict}(\textbf{x}_{t})] $ \hfill \% clustering modules
\STATE {\bfseries } \quad \quad \quad  $\textbf{b}_{a_t} \leftarrow \textbf{b}_{a_t}+r'\textbf{x}_{t}$ 
\STATE {\bfseries } \quad \quad \textbf{elif} \hfill \% ignore self-supervision signals
\STATE {\bfseries }  \quad \quad  \quad $\textbf{A}_{a_t} \leftarrow \textbf{A}_{a_t}+\textbf{x}_{t}\textbf{x}^{\top}_{t}$ 
\STATE {\bfseries } \quad \quad \textbf{end if}
\STATE {\bfseries } \quad \textbf{end if}
 \STATE {\bfseries }\textbf{end for}
 \end{algorithmic}
\end{algorithm}
\setlength{\textfloatsep}{0.35cm}
\setlength{\floatsep}{0.35cm}

\vspace{-0.2cm}

\subsection{Self-Supervision and Semi-Supervision Modules}
\label{sec:clustering}
\vspace{-0.1cm}

We construct our self-supervision modules given the cluster assumption of the semi-supervision problem: the points within the same cluster are more likely to share a label. As shown in many work in modern speaker diarization, clustering algorithms like GMM \cite{zajic2017speaker} and spectral clustering \cite{wang2018speaker} are especially powerful unsupervised modules, especially in their offline versions. Their online variants, however, often perform poorly \cite{zhang2019fully}. Nonetheless, we chose three popular clustering algorithms as the self-supervision: GMM, Kmeans and K-nearest neighbors (KNN), all in their online versions.

\vspace{-0.2cm}

\subsection{Complete Engine for Online Speaker Diarization}
\label{sec:engine}
\vspace{-0.1cm}

To adapt our BerlinUCB algorithm to the specific application of speaker recognition, we firdst define our actions. There are three major classes of actions: an arm ``New'' to denote that a new speaker is detected, an arm ``No Speaker'' to denote that no one is speaking, and N different arms ``User n'' to denote that user n is speaking. Table \ref{tab:routes} presents the reward assignment given four types of feedbacks. Note that we assume that when the agent correctly identifies the speaker (or no speaker), the user (as the feedback dispenser) should send no feedbacks to the system by doing nothing. In another word, in an ideal scenario when the agent does a perfect job by correctly identifying the speaker all the time, we are not necessary to be around to correct it anymore (i.e. truly feedback free). As we pointed out earlier, this could be a challenge earlier on, because other than implicitly approving the agent's choice, receiving no feedbacks could also mean the feedbacks are not revealed properly (e.g. the human oracle took a break). Furthermore, we noted that when ``No Speaker'' and ``User n'' arms are correctly identified, there is no feedback from us the human oracle (meaning that these arms would never have learned from a single positive reward if we don't use the ``None'' feedback iterations at all!). The semi-supervision by self-supervision step is exactly tailored for a scenario like this, where the lack of revealed positive reward for ``No Speaker'' and ``User n'' arms is compensated by the additional training of the reward mapping $\textbf{b}_{a_t}$ if context $\textbf{x}_t$ is assigned to the right arm.

To tackle the cold start problem, the agent grows it arms in the following fashion: the agent starts with two arms, ``No Speaker'' and ``New''; if it is actually a new speaker speaking, we have the following three conditions: (1) if ``New'' is chosen, the user approves this arm by giving it a positive reward (i.e. clicking on it) and the agent initializes a new arm called ``User $N$'' and update $N=N+1$ (where $N$ is the number of registered speakers at the moment); (2) if ``No Speaker'' is chosen, the user disapproves this arm by giving it a zero reward and clicking on the ``New'' instead), while the agent initializes a new arm; (3) if one of the user arms is chosen (e.g. ``User 5'' is chosen while in fact a new person is speaking), the agent copies the wrong user arm's parameters to initialize the new arm, since the voiceprint of the mistaken one might be beneficial to initialize the new user profile. In this way, we can transfer what has been learned for a similar context representations to the new arm.\footnote{Potential problems might occur if the number of users grows steadily through misclassifications. In future work, we will investigate possible branch pruning strategies and the processing of very sparse reward feedback.}

\begin{table}[tb]
\centering
\resizebox{\columnwidth}{!}{
 \begin{tabular}{ l | c | c | c | c }
  Feedback types & (+, +) & (+, -)  & (-, +) & \textit{None} \\ \hline
  New & $r=1$ & $r=0$ & - &\multirow{3}{*}{Alg. \ref{alg:berlinucb} Step 13} \\
  No Speaker & - & $r=0$ & $r=0$ & \\
  User n & - & $r=0$ & $r=0$ & \\
 \end{tabular}
 }
  \vspace{-0.3cm}
\caption{Possible algorithm routes given no feedbacks, or a feedback telling the agent that the correct label is $a*$. (+, +) means that the agent guessed it right by choosing the right arm; (+, -) means that the agent chose this arm incorrectly, since the correct one is another arm; (-, +) means that the agent didn't choose this arm, while it turned out to be the correct one. ``-'' marks scenarios not applicable.}
\label{tab:routes}
 \end{table}
 

\vspace{-0.2cm}
\section{Benchmark Description: \textit{MiniVox}}
\vspace{-0.1cm}

MiniVox is an automatic framework to transform any speaker-labelled dataset into continuous speech datastream with episodically revealed label feedbacks.
Since our online learning problem setting assumes learning the voiceprints without any previous training data at all, \textit{MiniVox}'s flexibility in length and configuration is especially important. As outlined in Figure \ref{fig:minivox}, \textit{MiniVox} has a straightforward data stream generation pipeline: given a pool of single-speaker-annotated utterances, randomly concatenate multiple pieces with a chosen number of speakers and a desired length. The reward stream is then sparsified with a parameter $p$ as the percentage of time a feedback is revealed. 

There are two scenarios that we can evaluate in MiniVox: if we assume there is an oracle, the online learning model is given the fixed number of the speakers in the stream; if we assume there is no oracle, the online learning model will start from zero speaker and then gradually discover and register new speakers for future identification and diarization.


\vspace{-0.1cm}
\section{Empirical Evaluation}
\vspace{-0.2cm}

\begin{figure*}[tb]
\centering
    \includegraphics[width=\linewidth]{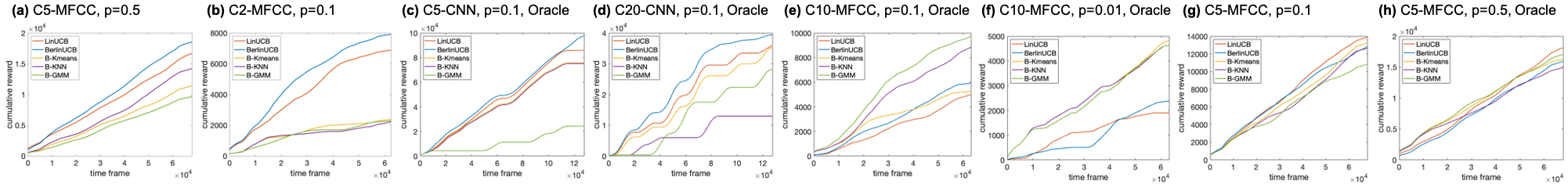}
\vspace{-0.7cm}
\caption{\textbf{Example reward curves} where \textbf{(a, b, c, d)} BerlinUCB is the best; \textbf{(e, f)} the self-supervision is the best; \textbf{(g, h)} LinUCB is the best.}\label{fig:results}
\vspace{-0.5cm}
\end{figure*}

\subsection{Experimental Setup and Metrics}
\vspace{-0.1cm}

We applied MiniVox on VoxCeleb \cite{Nagrani17} to generate three data streams with 5, 10 and 20 speakers to simulate real-world conversations. We extracted two types of features (more details in section \ref{sec:feat}) and evaluated it in two scenarios (with or without oracle). The reward streams are sparsified given a revealing probability of 0.5, 0.1 and 0.01. In summary, we evaluated our models in a combinatorial total of 3 speaker numbers $\times$ 3 reward revealing probabilities $\times$ 2 feature types $\times$ 2 test scenarios $=$ 36 online learning environments. The online learning timescale range from $\sim$12000 to $\sim$60000 timeframes, with a frame shift of 10 ms. For notation of a specific MiniVox, in this paper we would denote ``MiniVox C5-MFCC-60k'' as a MiniVox environment with 5 speakers ranging 60k time frames using MFCC as features. 

To evaluate performance in the above MiniVox environments, we reported Diarization Error Rates (DER), the standard introduced in the NIST Rich Transcription 2009 (RT-09). In addition, as a common metric in online learning literature, we also recorded the cumulative reward: at each frame, if the agent correctly predicts a given speaker, the reward is counted as +1 (regardless of whether the agent observes it or not). 

We compared 5 agents: The baseline, \textit{LinUCB} is the contextual bandit with extendable arms proposed in section \ref{sec:banditarm}. \textit{BerlinUCB} is our standard contextual bandit model designed for sparse feedbacks without the self-supervision modules. To test the effect of self-supervision, we introduced three clustering modules in BerlinUCB (Alg \ref{alg:berlinucb}, Step 15) denoted \textit{B-Kmeans}, \textit{B-KNN} (with K=5), and \textit{B-GMM}, whose clustering modules are randomly initialized and updated online. 

\vspace{-0.3cm}

\subsection{Feature Embeddings: MFCC and Neural Networks}
\label{sec:feat}
\vspace{-0.1cm}

We utilized two feature embeddings for our evaluation: MFCC \cite{hasan2004speaker} and a Convolutional Neural Network (CNN). We utilized the same CNN architecture as the VGG-M \cite{Chatfield14} used in VolCeleb evaluation \cite{Nagrani17}. It takes the spectrogram of an utterance as the input, and generate a feature vector of 1024 in layer fc8 (table 4 in \cite{Nagrani17} for details about this CNN). 


\textbf{Why don't we use more complicated embeddings?} Although more complicated embedding extraction modules such as i-vectors \cite{shum2013unsupervised} or d-vectors \cite{wang2018speaker} can improve diarization, they require extensive pretraining on big datasets, which is contradictory to our problem setting and beyond our scope. 

\textbf{Why do we still include a pretrained CNN embedding?} Indeed, if our end goal is to let the system learn from scratch without pretraining, why do we consider it in our evaluation?
The CNN model was trained for speaker verification task in VoxCeleb and we are curious about the relationship between a learned representation and our online learning agents. Despite this note, we are most interested in the performance given MFCC, because we aim to push the system to the extreme of not having pretraining of any type before deployment.

\begin{table}[t]
\begin{minipage}{\linewidth}
      \caption{Diarization Error Rate (\%) in MiniVox \textbf{without} Oracle
      }
      \vspace{-0.3cm}
      \label{tab:without} 
      \centering
      \resizebox{\linewidth}{!}{
 \begin{tabular}{ l | c | c | c | c | c | c | }
 &\multicolumn{3}{c}{MiniVox C5-MFCC-60k} \vline & \multicolumn{3}{c}{MiniVox C5-CNN-12k} \vline \\
  & $p=0.5$ & $p=0.1$ & $p=0.01$  & $p=0.5$ & $p=0.1$ & $p=0.01$ \\ \hline
BerlinUCB &  \textbf{71.81} & 80.03 & 82.38 & \textbf{17.42} & \textbf{32.03} & 65.16 \\
LinUCB &  74.74 & \textbf{78.71} & 79.30 & 17.81 & 32.73 & \textbf{58.98} \\ \hline
B-Kmeans &  82.82 & 79.15 & \textbf{77.39} & 28.83 & 63.67 & 82.58 \\
B-KNN &  78.71 & 80.62 & \textbf{77.39} & 28.36 & 82.58 & 82.58 \\
B-GMM & 85.32 & 83.41 & 87.67 & 99.61 & 99.61 &  99.69 \\ \hline
 \end{tabular}
 }
 \end{minipage}

\begin{minipage}{\linewidth}
      \centering
      \resizebox{1\linewidth}{!}{
 \begin{tabular}{ l | c | c | c | c | c | c | }
 &\multicolumn{3}{c}{MiniVox C10-MFCC-60k} \vline & \multicolumn{3}{c}{MiniVox C10-CNN-12k} \vline \\
  & $p=0.5$ & $p=0.1$ & $p=0.01$  & $p=0.5$ & $p=0.1$ & $p=0.01$ \\ \hline
BerlinUCB & \textbf{82.46} & \textbf{85.31} & \textbf{89.26}  & \textbf{42.77} & \textbf{57.41} & \textbf{74.02} \\
LinUCB & 84.36 & 86.73 & 93.36  & 49.55 & 68.57 & 81.16  \\ \hline
B-Kmeans & 91.15 & 92.58 & 96.68 & 60.89 & 70.89 & 99.55  \\
B-KNN & 89.73 & 90.05 & 96.68 & 60.89 & 82.05 & 99.55 \\
B-GMM & 90.21 & 94.63 & 98.42  & 99.20 & 93.57 & 99.64  \\ \hline
 \end{tabular}
 }
 \end{minipage}

\begin{minipage}{\linewidth}
      \centering
      \resizebox{1\linewidth}{!}{
 \begin{tabular}{ l | c | c | c | c | c | c | }
 &\multicolumn{3}{c}{MiniVox C20-MFCC-60k} \vline & \multicolumn{3}{c}{MiniVox C20-CNN-12k} \vline \\
  & $p=0.5$ & $p=0.1$ & $p=0.01$  & $p=0.5$ & $p=0.1$ & $p=0.01$ \\ \hline
BerlinUCB & \textbf{88.62} & \textbf{87.02} & 92.79 & \textbf{41.72} & \textbf{59.06} & 83.28 \\
LinUCB & 91.35 & 88.94 & \textbf{88.46} & 51.56 & 83.52 & \textbf{74.84} \\ \hline
B-Kmeans & 95.19 & 95.99 & 96.96 & 72.03 & 75.31 & 99.53 \\
B-KNN & 93.43 & 95.99 & 96.79 & 72.03 & 74.06 & 99.53 \\
B-GMM & 92.79 & 96.31 & 97.76 & 87.73 & 81.09 & 83.28 \\ \hline
 \end{tabular}
 }
 \end{minipage}
       \vspace{-0.2cm}
 \end{table}

\begin{table}[t]
\begin{minipage}{\linewidth}
      \caption{Diarization Error Rate (\%) in MiniVox \textbf{with} Oracle
      }
            \vspace{-0.3cm}
      \label{tab:with} 
      \centering
      \resizebox{\linewidth}{!}{
 \begin{tabular}{ l | c | c | c | c | c | c | }
 &\multicolumn{3}{c}{MiniVox C5-MFCC-60k} \vline & \multicolumn{3}{c}{MiniVox C5-CNN-12k} \vline \\
  & $p=0.5$ & $p=0.1$ & $p=0.01$  & $p=0.5$ & $p=0.1$ & $p=0.01$ \\ \hline
BerlinUCB & 74.89 & 77.24 & 86.93 & \textbf{17.27} & \textbf{22.19} & 66.02 \\
LinUCB & \textbf{72.83} & 78.12 & \textbf{76.80} & 17.73 & 32.73 & \textbf{58.98} \\ \hline
B-Kmeans & 75.33 & 78.27 & 83.11 & 20.55 & 40.70 & \textbf{58.98}  \\
B-KNN & 77.39 & 77.97 & 83.99 & 20.47 & 41.33 & \textbf{58.98}  \\
B-GMM & 74.16 & \textbf{76.21} & 77.24 & 52.58 & 81.02 & \textbf{58.98} \\ \hline
 \end{tabular}
 }
 \end{minipage}

\begin{minipage}{\linewidth}
      \centering
      \resizebox{1\linewidth}{!}{
 \begin{tabular}{ l | c | c | c | c | c | c | }
 &\multicolumn{3}{c}{MiniVox C10-MFCC-60k} \vline & \multicolumn{3}{c}{MiniVox C10-CNN-12k} \vline \\
  & $p=0.5$ & $p=0.1$ & $p=0.01$  & $p=0.5$ & $p=0.1$ & $p=0.01$ \\ \hline
BerlinUCB & 88.31 & \textbf{90.21} & 95.89 & \textbf{45.18} & \textbf{65.27} & 79.38 \\
LinUCB & \textbf{84.99} & 91.63 & 97.00 & 50.00 & 72.14 & \textbf{65.18} \\ \hline
B-Kmeans & 87.84 & 91.47 & \textbf{91.94} & 50.27 & 72.50 & 72.32 \\
B-KNN & 86.73 & 85.78 & 92.58 & 49.64 & 72.14 & 77.77 \\
B-GMM & 88.94 & 84.52 & 92.58 & 76.52 & 71.88 & 69.46 \\ \hline
 \end{tabular}
 }
 \end{minipage}
        

\begin{minipage}{\linewidth}
      \centering
      \resizebox{1\linewidth}{!}{
 \begin{tabular}{ l | c | c | c | c | c | c | }
 &\multicolumn{3}{c}{MiniVox C20-MFCC-60k} \vline & \multicolumn{3}{c}{MiniVox C20-CNN-12k} \vline \\
  & $p=0.5$ & $p=0.1$ & $p=0.01$  & $p=0.5$ & $p=0.1$ & $p=0.01$ \\ \hline
BerlinUCB & 92.31 & 94.55 & 96.31 & 58.75 & \textbf{68.98} & 88.83 \\
LinUCB & \textbf{89.10} & 93.43 & \textbf{95.67} & \textbf{53.44} & 70.47 & \textbf{83.44} \\ \hline
B-Kmeans & 92.95 & 95.67 & 96.96 & 55.16 & 70.86 & 94.06 \\
B-KNN & 91.83 & 92.47 & 97.44 & 54.30 & 89.84 & 96.72 \\
B-GMM & 95.19 & \textbf{91.99} & 97.44 & 86.48 & 77.97 & 96.64 \\ \hline
 \end{tabular}
 }
 \end{minipage}
       \vspace{-0.3cm}
 \end{table}


\vspace{-0.2cm}

\subsection{Results}
\vspace{-0.1cm}

Given MFCC features without pretraining, our online learning agent demonstrated a robust performance (Figure \ref{fig:results}a,b,c,d): in most cases, it significantly outperformed the baseline. \footnote{We wish to note the overall high diarization error rate in all MFCC benchmarks: it is important to keep in mind that the bandit feedback (correct or incorrect classification) makes the \textit{online} speaker diarization problem significantly more challenging, as compared to the standard supervised learning in \textit{offline} speaker diarization, since the true label is never revealed in bandit setting unless the classification is correct. Thus, the diarization error rate in a bandit online setting is expected to be much higher than in the supervised learning setting, which is not due to inferiority of bandit decision making algorithm versus other classifiers, but due to increased problem difficulty.}

\textbf{Learning without Oracle} (Table \ref{tab:without}).
In both the MFCC and CNN MiniVox environments, we observed that BerlinUCB and its variants outperform the baseline most of the time. The discrepancy of performance of the MFCC and CNN environments can be explained by the innate difficulties of the two tasks: while the CNN embeddings are already well separated because they were pretrained with contrastive loss \cite{Nagrani17}, in MFCC environments our online learning models need to learn from scratch both how to cluster and how to map the reward with the features, while maintaining a good balance between exploitation and exploration. 


\textbf{Learning with Oracle} (Table \ref{tab:with}).
Given the number of speakers, the online clustering modules appears to be more effective. However, the behaviors vary: we observed that B-GMM performed the poorest in the oracle-free environments, but performed the best in some environments with oracle; we also noted that despite the consistent best model in many oracle-free environments, the standard BerlinUCB was surpassed by the baseline and its self-supervised variants in a few MFCC cases with oracle; in certain challenging cases where the reward is sparsely revealed (p=0.01 or 0.1), the self-supervised variants improve the performance of BerlinUCB.


\textbf{Is self-supervision useful?} To our surprise, our benchmark results suggested that for most cases, the proposed self-supervision modules didn't improve upon our proposed contextual bandit model. Only in specific conditions (e.g. MiniVox C20-MFCC-60k p=0.1 with Oracle), the self-supervised contextual bandits outperformed both the standard BerlinUCB and the baseline. Further investigation into the reward curve revealed more complicated interactions between the self-supervision modules with the online learning modules (the contextual bandit): as shown in Figure \ref{fig:results}e,f, B-GMM and B-KNN built upon the effective reward mapping from their BerlinUCB backbone, and benefited from the unlabelled data points to yield a fairly good performance.

\vspace{-0.2cm}
\section{Conclusion and outlook}
\vspace{-0.2cm}

We considered the novel problem of online learning speaker diarization.
We formulated the practical task as an interactive system that episodically receives sparse bandit feedback from users. During unlabelled episodes, we proposed to learn from pseudo-feedback generated by self-supervised modules enabled by clustering. We provided a benchmark to evaluate this task, and demonstrated an empirical merit of the proposed methods over standard online learning algorithm. 
Ongoing work include extending the online learning framework in both extraction and clustering modules, branch management (e.g. routing \cite{lin2018contextual}) and self-supervision with graph methods.
\clearpage
\bibliographystyle{IEEEbib}
\bibliography{main}

\end{document}